\documentclass{article}
\usepackage{url}
\usepackage{graphicx} 
\usepackage{amsmath}

\usepackage{algorithm}
\usepackage{algpseudocode}

\usepackage{courier}
\usepackage{booktabs}
\usepackage{times}
\usepackage{helvet}

\title{Explore the Loss space with Hill-ADAM}

\author{Meenakshi Manikandan \\
Department of Computer Science and Engineering \\
UC San Diego \\
2mmanikan@gmail.com \\
\\
Dr. Leilani Gilpin \\
Department of Computer Science and Engineering \\
UC Santa Cruz \\
lgilpin@ucsc.edu
}
\date{October 3, 2025}

\begin{document}

\maketitle
\section{Abstract:}
This paper introduces Hill-ADAM, an optimizer with its focus towards escaping local minima in prescribed loss landscapes, to find the global minimum. Hill-ADAM escapes minima by deterministically exploring the state space, eliminating uncertainty from random gradient updates in stochastic algorithms while seldom converging at the first minimum that visits. In the paper, we first derive an analytical approximation of the ADAM Optimizer’s step size at a particular model state, and from there define the primary condition determining ADAM’s limitations in escaping local minima. The proposed optimizer algorithm Hill-ADAM alternates between error minimization and maximization, maximizing to escape the local minimum and minimizing again afterward. This alternation provides an overall exploration throughout the loss space, allowing the deduction of the global minimum's state. Hill-ADAM was tested with 5 loss functions and 12 amber-saturated to cooler-shade image color correction instances.

\maketitle

\section{Introduction:}
The ability for gradient-based optimization algorithms to escape local optima has been questioned for decades. Stochastic Gradient Descent/Ascent (SGD), the backbone of several of the optimization algorithms today, updates the model’s weights based on the gradient values obtained at each iteration of training. SGD operates in a manner in that once a minimum is reached, the algorithm seldom updates the model state. The model stops learning, in other words. Such an optimization method proves well for numerous applications in machine learning, with the condition that the respective loss function has a single, global minimum. In situations where the loss space contains multiple minima (local and global minima), SGD may stop optimizing when it reaches a local minimum rather than the global one. In such cases, the optimizer must escape the local minimum, which requires altering the extent to which the model's state is altered each time. A proposed solution is using the expectation of all the encountered gradients, rather than the raw gradient at each iteration, to escape the local minimum. This method, called momentum, allows the optimizer to continue updating the model state despite having reached a local minimum. The ADAM optimizer combined the idea of momentum with scaling the step size based on the gradient's variance~\cite{adam2014method}. The aim was to increase the step size during earlier, more uncertain stages of training, which allowed the optimizer to further push past the local minima. 

The use of expectation and variance in the gradient step calculation is ADAM's key factor to escaping local extrema, due to factoring in the previous gradient values. When the optimizer reaches a local minimum, rather than converging, it continues to travel along the loss space in the direction of the previously encountered gradients (in the gradient expectation and variance). Depending on the nature of the loss function, this gain in momentum by the optimization algorithm may surpass the local minimum and converge at the global minimum. However, there are several scenarios where ADAM oscillates around and eventually converges towards the local minima, which can happen when the calculated step changes direction based on the gradient expectation and variance.   

Due to ADAM's situational oscillation tendency, ADAM is prone to trapping itself in local extrema. ADAM’s exact success condition can be derived (see Section III). We propose a new optimizer: Hill-ADAM. The optimizer relying on gradient and momentum and variance alone. It maximizes the loss until the local minimum encountered before has been escaped, in which case the optimizer minimizes the loss once more. The optimizer allows the model to travel through different minima and maxima in the objective function over time. As training finishes, the model state corresponding to the least minima is retained.

\maketitle

\section{Related Works:}
\(\)

There are several related and common optimization strategies, such as Root Mean Square Propagation~\cite{hinton2012rmsprop}, Nesterov Adaptive Moment Estimation~\cite{dozat2016incorporating}, Rectified Adaptive Moment Estimation~\cite{liu2019variance}, and Simulated Annealing~\cite{kirkpatrick1983optimization}. Root Mean Square Propagation (RMSprop) is similar to ADAM in optimization strategy, but calculates the step size based on the gradient itself rather than the gradient's expectation. Using the current gradient can significantly vary the step size at different points in the curve without previous gradients included in the calculation. This may decrease training reliability, making it uncertain whether the optimizer will converge at the global minimum. 

Nesterov Adaptive Moment Estimation (NADAM) uses hyperparameter scheduling to alter the gradient expectation, more specifically, it corrects the expectation's initialization bias. The algorithm incorporates aspects of the Nesterov's accelerated gradient to decrease the convergence time that it takes to reach a solution. The algorithm is extremely fast, though it is not mathematically guaranteed that the global minima could be reached. Similarly, Rectified Adaptive Moment Estimation (RADAM) focuses on convergence quality. This is done by adding a correction term to the variance, preventing the variance from skyrocketing during the beginning of training. Though like NADAM global minima convergence is not the mathematical focus, raising uncertainty on its ability to converge towards the global minima.

Simulated Annealing, however, has been a common strategy towards escaping local minima. It adds randomization to step size, allowing the algorithm to explore the loss space rather than follow the traditional greedy approach of attempting minimization at each step. During the early stages of training, statistically more random steps are taken in the loss space with the goal of searching for the global minimum, rather than settling at a local minimum it would reach initially. However, Simulated Annealing's ability to reach the global minima is asymptotically optimal, as we are not sure whether the random steps helped the model converge overall, or if the steps were enough to escape the local minima the algorithm converged at.

\maketitle
\section{Background: }
The ADAM optimizer's step size is shown in Equation 1. \(\lambda\) is the learning rate. \(E[g]\) is the gradient expectation, or the moving average of gradients (momentum), and \(E[g^2]\) is the moving average of squared gradients (variance). \(\epsilon\) is an extremely small value, often set to the value 1e-8, and is used to ensure the stability of the optimizer in case \(E[g^2]\) equals zero. \(g_{t}\) is the gradient at step \(t\), and  \(g_{avg}\) is the moving average of the gradients (before the addition of the new gradient value). \( (g_{avg})_t \) is the moving average including the $t$'th gradient value. \(\Delta x\) represents the step size between the input vectors.

\begin{equation}
       \lambda * \frac{E[g]}{\sqrt{E[g^2]} + \epsilon}
\end{equation}

The moving average is a defining property of the optimization algorithm for ADAM. Refer to Equations 2 and 3 for the definitions of momentum and moving average.

\begin{equation}
      E[g] = \beta g_{avg}  + (1-\beta)g_{t}
\end{equation}

\begin{equation}
      (g_{avg})_{t} = \begin{cases}
        g_{0} & t = 0 \\ 
        \beta (g_{avg})_{t-1}  + (1-\beta)g_{t} & t>0 
         
      \end{cases}
      \label{eq3}
\end{equation}

\maketitle
\section{ADAM Optimizer Convergence Condition:}
We start by expanding the gradient expectation into a series form. The function $f$ represents the prescribed loss space. The $w$ coefficients represent the corresponding weight given each of $f$'s gradients. $x_0$ represents the initial input into the function and \(\Delta x\)(small value) represents the distance between each point in the loss space where the gradients are calculated to obtain the expectation. So, the resulting values that are input into function $f$ values represent the input vectors used to obtain gradients necessary to compute $E[g]$. Also, $n$ is the number of gradients included in the calculation.

The summation allows us to use the gradient at multiple points of the loss curve to obtain the gradient expectation. The gradient values are taken at several equidistant locations, which are separated by the constant value \(\Delta x\). This creates a collection of \(x\) values at which we take the gradients for $E[g]$, which can also be represented in the learning space as a trace along the loss function. 

\begin{equation}
    E[g] = \sum^N_{n=0} w_{n} * f'(x_{0}+n\Delta x), N=\frac{|x_{N} - x_{0}|}{\Delta x}
\end{equation}

From Equation 4, we can separate the first term from the summation to obtain Equation 5.

\begin{equation}
    E[g] = w_{0}f'(x_{0}) + \sum^N_{n=1} w_{n} * f'(x_{0}+n\Delta x)
\end{equation}

By the definition of the moving average (momentum), it is possible to obtain the weight values of each gradient term used to calculate $E[g]$ by expanding its definition. We start by defining the gradient expectation, which is represented by the moving average shown in Equation 2. Combining Equations 2 and 3 yields Equation 6.

\begin{equation}
    E[g] = \beta(\beta( (g_{avg})_{t-2} ) + (1-\beta)g_{t-1}) + (1-\beta)g_{t}
\end{equation}

Equation 6 is accurate for when three consecutive gradient values are used in the expectation calculation. With the goal of obtaining the weight values of all gradients included in the gradient expectation calculation (ie. gradients at each of the $t$ steps), we then expand Equation 6 to $t$ gradient values. To do so, the general representation of the gradient expectation can be represented as Equation 7.

\begin{equation}
    E[g] = \beta(\beta(\beta(...)+(1-\beta)g_{t-2}) + (1-\beta)g_{t-1}) + (1-\beta)g_{t}
\end{equation}

By simplifying Equation 7, we can obtain the weight values for each of the gradients to obtain Equation 8.

\begin{equation}
    E[g] = \beta^t g_0 + \beta^{t-1}(1-\beta)g_1 + ... + (1-\beta)g_t
\end{equation}

As each gradient at each of the $t$ steps have a clear coefficient (ex. $\beta^t$ for $g_0$), we can assign these very values to the weights of the gradients used in calculating $E[g]$, for a given number of gradients G.

\begin{equation}
      w_{t} = \begin{cases}
        \beta^G & t = 0 \\ 
        \beta^{G-t}(1-\beta) & 0<t\leq G 
      \end{cases}
\end{equation}

When substituting Equation 9 into Equation 5, the result is Equation 10.

\begin{equation}
    E[g] = \beta^N f'(x_0)  +  \sum^N_{n=1} (\beta^{N-n} (1-\beta)f'(x_0+n\Delta x))
\end{equation}

The variance of the gradients can be obtained using the same approach, as shown in Equation 11.

\begin{equation}
    E[g^2] = \beta^N f'(x_0)^2  +  \sum^N_{n=1} (\beta^{N-n} (1-\beta)f'(x_0+n\Delta x)^2)
\end{equation}

Substituting Equation 10 and Equation 11 into Equation 1 results in the ADAM Optimizer’s step size approximation in Equation 12.

\begin{equation}
    (S_x)_n = \lambda * \frac{ \beta^N f'(x_0)  +  \sum^N_{n=1} (\beta^{N-n} (1-\beta)f'(x_0+n\Delta x)) }{ \sqrt{ \beta^N f'(x_0)^2  +  \sum^N_{n=1} (\beta^{N-n} (1-\beta)f'(x_0+n\Delta x)^2)  } + \epsilon  }
\end{equation}

The ADAM optimizer stops updating when the step size reaches zero, and in most cases changes sign. It is important to note that the step size in said situation is not exactly zero, but approximates it, which is the case in most prescribed loss space-based machine learning scenarios. So, the official condition in which the ADAM optimizer escapes the local extrema is shown in Equation 13.

\begin{equation}
      \begin{cases}  
        \lambda * \frac{ \beta^N f'(x_0)  +  \sum^N_{n=1} (\beta^{N-n} (1-\beta)f'(x_0+n\Delta x)) }{ \sqrt{ \beta^N f'(x_0)^2  +  \sum^N_{n=1} (\beta^{N-n} (1-\beta)f'(x_0+n\Delta x)^2)  } + \epsilon  } \approx 0 & x_N - x_{extrema} \geq 0 \\ \\
        
        \lambda * \frac{ \beta^N f'(x_0)  +  \sum^N_{n=1} (\beta^{N-n} (1-\beta)f'(x_0-n\Delta x)) }{ \sqrt{ \beta^N f'(x_0)^2  +  \sum^N_{n=1} (\beta^{N-n} (1-\beta)f'(x_0-n\Delta x)^2)  } + \epsilon  } \approx 0 & x_N - x_{extrema} < 0
        
      \end{cases}
\end{equation}

The numerator must either approach zero or the denominator must approach infinity for the former condition to hold. The extrema escape condition for ADAM is quite specific, and definitely seldom applies to all prescribed loss functions that are to be encountered in machine learning applications. 

Equation 13 serves as an \textit{approximation} of the parameter (x-value) in which the ADAM optimizer fails. The main reason for this is that when the gradient expectation is calculated, the sequence of encountered gradients are equidistant from each other, which is not always be the case. The properties of the curve (notably the absolute value of the gradient at each step) affect the distances between the points in which the gradient is taken. For example, if the optimizer starts at parameters in which the loss function derivative was extremely large (extremely steep curve), the steps taken would be extremely large and more variant in distance between states (in which the gradient is taken). Equation 13 can, when slightly modified, account for this varying difference in steps between each location on the loss curve. Still, even the approximation conveyed in Equation 13 provides insight on how \textit{an infinite number} of traces along the loss function will fare in reaching the global minimum through ADAM optimization.

\maketitle
\section{Hill-ADAM Optimization Algorithm:}

\begin{algorithm}[H]
\caption{Hill-ADAM -- One Training Step}\label{alg:cap}
\begin{algorithmic}[1]

\State $Required: \lambda(learning\hspace{0.1cm} rate), (\beta_1, \beta_2) (parameters\hspace{0.1cm} for\hspace{0.1cm} momentum\hspace{0.1cm} and\hspace{0.1cm} variance\hspace{0.1cm} calculation)$
\State $Required: \delta (threshold\hspace{0.1cm} in\hspace{0.1cm} which\hspace{0.1cm} to\hspace{0.1cm} switch\hspace{0.1cm} current\hspace{0.1cm} direction), \gamma (dead\hspace{0.1cm} end\hspace{0.1cm} threshold)$
\State
\State $set\_to\_zero\left[E\left[g\right],E\left[g^2\right], step\right]$
\State $set\_to\_one\left[not\_deadend\right]$
\State $current\_direction=minimize, smallest\_loss=\infty$

\State
  
\If{$abs(new\_loss-previous\_loss)<\delta$}
    \State $current\_direction \gets toggle[current\_direction]$
    \State $reset[previous\_loss, new\_loss]$
\EndIf

\State

\If{$new\_loss>\gamma \hspace{0.1cm}$}
    \State $current\_direction \gets minimize$
    
    \State $unmark\_as\_dead\_end$
    \State $reset[previous\_loss, new\_loss]$
    
    \Else
        \State $mark\_as\_dead\_end$
\EndIf

\State

\ForAll{$param \gets model\_parameters$}
    \If{$direction\_has\_changed \hspace{0.1cm} $}
        \State $set\_to\_zero\left[E\left[g\right],E\left[g^2\right], step\right]$
    \EndIf

    \State $step \gets (step+1)$

    \State $grad \gets gradient\left[param\right]$
    \State $E\left[g\right] \gets \beta_1 E\left[g\right] + \left(1-\beta_1\right)grad$
    \State $E\left[g^2\right] \gets \beta_2 E\left[g^2\right] + \left(1-\beta_2\right)grad^2$
    \State $E\left[g\right] \gets E\left[g\right]/(1-(\beta_1)^{step})$
    \State $E\left[g^2\right] \gets E\left[g^2\right]/(1-(\beta_2)^{step})$

    \If{$minimize$}
        \State $param \gets param-\lambda(E\left[g\right]/\sqrt{(E\left[g^2\right] + \epsilon)}) $
        \Else
        \State $param \gets param+\lambda(E\left[g\right]/\sqrt{(E\left[g^2\right] + \epsilon)}) $
    \EndIf
    
\EndFor

\State

\State $update[previous\_loss, minimum\_loss, minimum\_model\_state]$

\end{algorithmic}
\end{algorithm}

The objective is to minimize the loss. The algorithm begins by minimizing the loss (as per the supplied prescribed loss function) \textit{ as per ADAM’s algorithm}. ADAM calculates the step size using the step number, learning rate \(\lambda\), epsilon \(\epsilon\), gradient expectation (\(E[g]\)), gradient variance (\(E[g]^2\)), along with moving average parameters \(\beta_1\) and \(\beta_2\). This takes place until the algorithm reaches the point where the step size is near zero. 

The step size is considered zero when the difference between the newly-acquired loss and previously found loss is below \(\delta\). We refer to such a model state as a critical state. Once a critical state is reached, the algorithm, for the sake of being conservative, assumes that the condition in Equation 13 is \textit{not} satisfied. The algorithm essentially operates under the assumption that it was in the process of climbing out of the local minima and is trapped due to the zero step size.

As a result, the optimizer shifts its focus temporarily to escaping the local minima that it is stuck in. So, the optimizer aims to maximize the loss until the loss difference is approximately zero once again. This signifies that the optimizer is now trapped in a local \textit{maximum}. Though slightly less intuitive, this scenario is the opposite problem to being stuck at a local \textit{minimum}. It is important to note that the step size, expectation of gradient, and gradient variance have been \textit{reset to zero} when maximization begins. This is done so that previous gradients do not effect convergence. 

There are also cases in which the model reaches a \textit{dead end}, in which the error may increase towards an infinite value. The limit of the prescribed function, when approaching the direction that the algorithm is moving in, is $\inf$ in other words. In such cases, we set a threshold \(\gamma\) for the algorithm at the beginning of training. If the loss value goes any higher than  \(\gamma\), that means a dead end has been reached and the algorithm must begin minimizing immediately.

The goal is then shifted back to the original plan of minimizing the loss function, and the alternation process repeats. Hill-ADAM has an additional feature in that it stores the model’s best encountered state. The optimizer updates and stores the lowest loss so far as well as its corresponding state. This is so that the best state can be reused at the end of training. Hill-ADAM explores the bumpy loss space in search of the global minimum.

\maketitle
\section{Experiments:}
\subsection{Experiment 1: Polynomial Loss Functions}

Five polynomials of differing orders were used as loss functions, each minimized by both Hill-ADAM and ADAM optimizers. This was done to compare Hill-ADAM's optimization capability (with prescribed loss functions) with ADAM’s. The five loss function polynomials used can be found in Table 1.

We design an artificial neural network as shown in Figure 1 with the task of predicting the state, or $x$-value, that yields the global minimum value of each loss function. The neural network has an input layer of six nodes that is fully connected to a hidden layer of four nodes, which is then connected to another hidden layer of three nodes. The latter is connected to the output layer which contains one node. The output from this node (the parameter) is then substituted into the loss function to evaluate the loss, which in turn trains the neural network. 

There are two variables to consider when it comes to the neural network training: the input to the network, and the weights of the network that update with each step. In this experiment, we keep the input to the network constant throughout training. This is because the experiment's goal is to analyze the specific optimization process of the network's weights. 

For all training instances, the ADAM and Hill-ADAM optimizer had a learning rate of 0.01. For Hill ADAM, the \(\delta\)-value, or the difference in losses before switching optimization goal to maximize/minimize, is set to 0.0001. 15000 training steps were used in all cases. 15000 training steps were chosen to give Hill-ADAM enough training steps to thoroughly explore the defined loss space. It is also worth noting that each optimizer-loss function combination was trained 15 times to ensure thoroughness in data collection. The mean of the resulting minima from each experiment is shown in Table 1. The learning rate of 0.01 was chosen to ensure quick convergence while avoiding model instability from extreme overshooting. The loss difference threshold of 0.0001 was chosen to allow for any slight changes in error as a stepsize will seldom ever be perfectly 0, which is our theoretical assumption for the beginning of local minima convergence as shown in Algorithm 1.

\begin{center}
\includegraphics[width = 8 cm, height = 5cm]{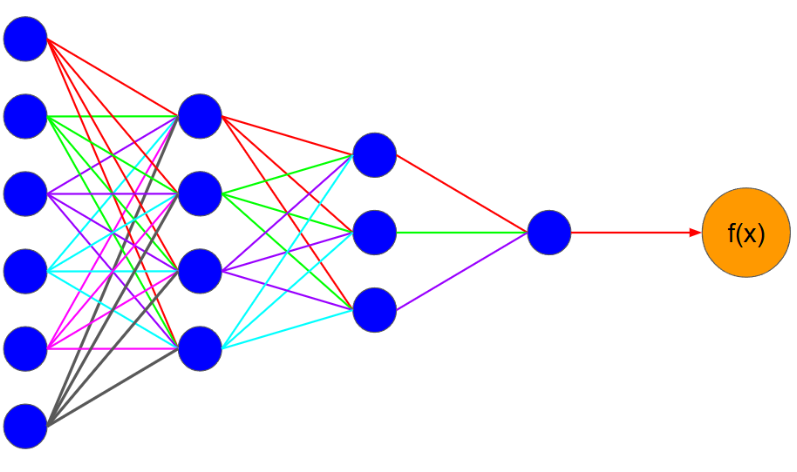}
\end{center}

\textbf{
Figure 1: Artificial neural network (nodes in dark blue) architecture used in first experiment. Network predicts parameter used to minimize the loss function (in orange).} \\

\begin{tabular}{c|c|c|}
    \hline
     & ADAM & Hill-ADAM  \\
     \hline
     
     \(x^2 + 4\) &  4.000 & 4.0000 \\
    \hline
    
    \(40 * (x^6+x^5+3x^3+4x^2+x) + 8\) &  4.4237 & 3.3475 \\

    \hline
   \( 4x^8-76x^7+605x^6-2614.01x^5+\) \\ \(6629x^4-9940x^3+8404x^2-3552x+576   \) &  -0.4837 & -10.4340 \\

  \hline 

   \( 0.1x^10-3x^9+39.5x^8-300x^7+1452.3x^6-\) \\ \(4671x^5+10080.5x^4-14370x^3+\) \\ \(12907.6x^2-6576.3x+1440   \) &  -0.5425 & -1.3811 \\

    \hline
   \( x^4 - 6x^3 +12x^2-10x+5  \) &  0.3125 & 0.3125

\end{tabular}
\\

\textbf{
Table 1: The table entries represent the mean of the resulting minima for each optimizer-loss function combination (each training instance).
} \\

From Table 1, we can conclude that Hill-ADAM reaches a consistently lower minimum, or equal minimum, to ADAM. 

\includegraphics[width = 7cm, height = 7cm]{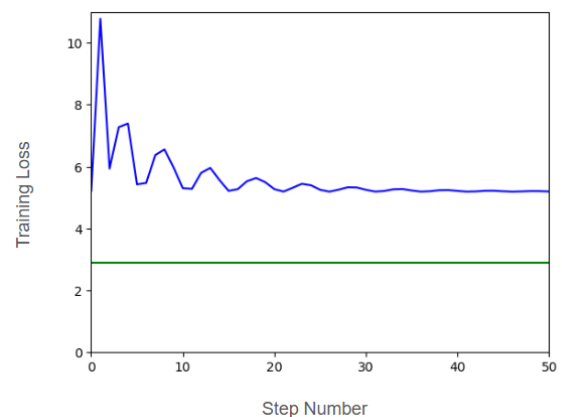}
\includegraphics[width = 7.1cm, height = 7.1cm]{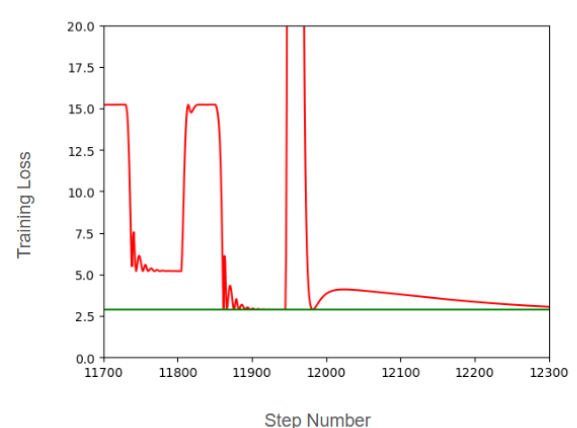} \\
\begin{center}

\textbf{
Figure 2: ANN in Figure 1 trained with ADAM (left) and Hill-ADAM (right), with the loss function being the sixth order polynomial function denoted in Table 1. Green line represents the loss function's theoretical global minimum. Portions of graphs chosen solely for the purpose of demonstrating Hill-Adam's trajectory (does not necessarily represent first timestep in which global minimum is found). }
\end{center}

\includegraphics[width = 7.5cm, height = 7.5cm]{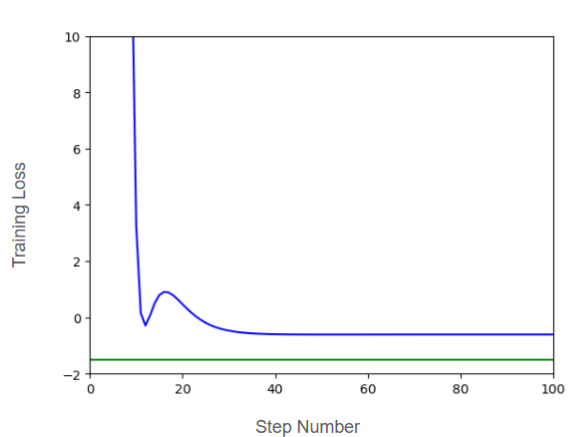}
\includegraphics[width = 6.7cm, height = 6.7cm]{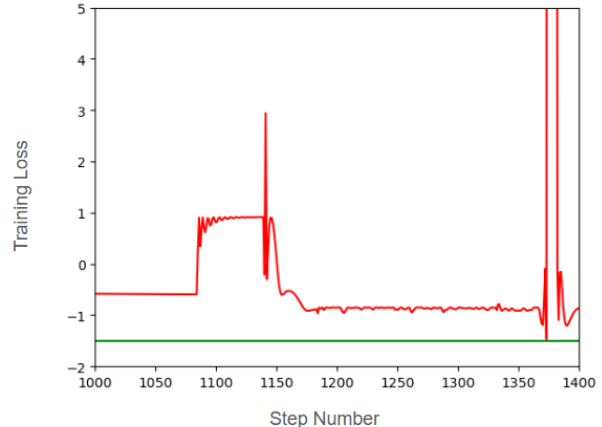}
\begin{center}
\textbf{
Figure 3: ANN in Figure 1 trained with ADAM (left) and Hill-ADAM (right). The loss function used is the tenth order polynomial denoted in Table 1. Portions of graphs chosen solely for the purpose of demonstrating Hill-Adam's trajectory. }
\end{center}

\subsection{Experiment 2: Application of Hill-ADAM to Color Correction}

In this experiment, we trained another Neural Network (different number of layers/neurons from the first experiment) to find a vector that transforms the color palette of a given source image to match that of a given target image. For each pair of images, we trained the model twice: once using ADAM and the other time using Hill-ADAM. We found that Hill-ADAM consistently outperformed ADAM.

Each color channel (RGB) in both the source and target images are viewed as statistical distributions~\cite{reinhard2002color}. The original paper's approach first takes the source and target images, and calculates the mean value and standard deviation for each RGB channel for each image. These values are then applied to the source image to match the target image's distribution.

For the purposes of this paper and optimizer usage, we use a modified approach to color correction. Similar to the original paper we treat the images as statistical distributions, but we only calculate the means for each channel in the source and target images. We omit the standard deviations for the sake of experimental simplicity (though the idea is the same even if the standard deviations are added). The means are then used to construct the \textit{prescribed loss function}, where the network's learning objective is to match the source mean to the target mean for each RGB channel. This is done by learning each channel's gain (coefficient of the source). These gains construct the \textit{vector} that transforms the color palette of the source image.

The prescribed loss function representing the simple mean approximation has only one true global minimum, so we explore the crucial concept of regularization (i.e. loss function modification) in color correction. In the case that the target image is intended to be a guide (not a benchmark) for our source image's new color palette, and that the truly necessary target image is not present, additional constraints may be needed for the color correction. The use of a neural network to learn gains is even more crucial in this context, as it is more straightforward to \textit{generalize} these conditions by adding them to a prescribed loss function (rather than using trial and error to modify gain as required with the original paper's method of color correction). This addition to the loss function can create several local minima. Such a scenario, with the presence of potentially many local minima, will demonstrate the extent of the difference in performance between ADAM and Hill-ADAM.

For this experiment, we aim to color-correct images saturated with yellow/orange shades to cooler, realistic shades. In other words, we want to avoid making the image too blue-saturated. The target images that we provide, however, are saturated with blue, which requires that we add regularization in addition to distribution approximation. The learning rate used for both ADAM and Hill-ADAM were 1e-2, and the number of training steps used for both were 1000.

\begin{tabular}{c|c|c|c|c}
    \hline
     & Seed 5343 & Seed 100 & Seed 2534 & Seed 3956  \\
     \hline
     ADAM &  2.9471 & 2.9471 & 2.9471 & 2.9471 \\
    \hline
    RMSprop & 2.6051 & 2.5917 & 3.6865 & 2.6482 \\
    \hline
    NADAM & 2.5308 & 2.5310 & 2.5321 & 2.5372 \\
    \hline
    Hill-ADAM (Ours) & 2.5305 & 2.5307 & 2.5306 & 2.5697 \\

\end{tabular} \\

\textbf{Table 2: The table represents the minimum reached by each of the optimizers, given the seed number as well.} \\

\includegraphics[width = 6.9cm, height = 6.9cm]{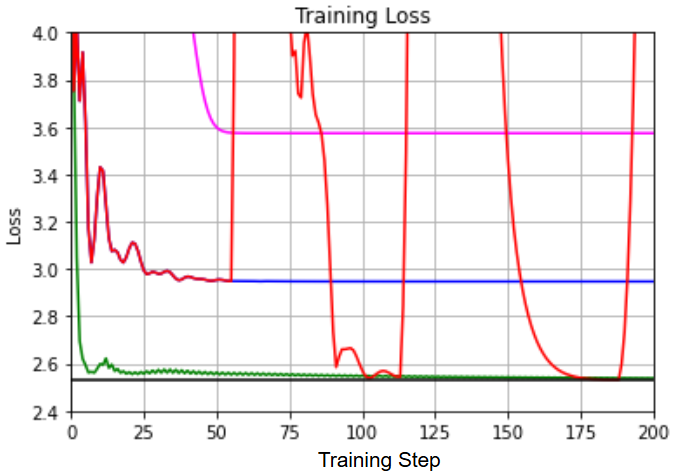} 
\includegraphics[width = 7cm, height = 7cm]{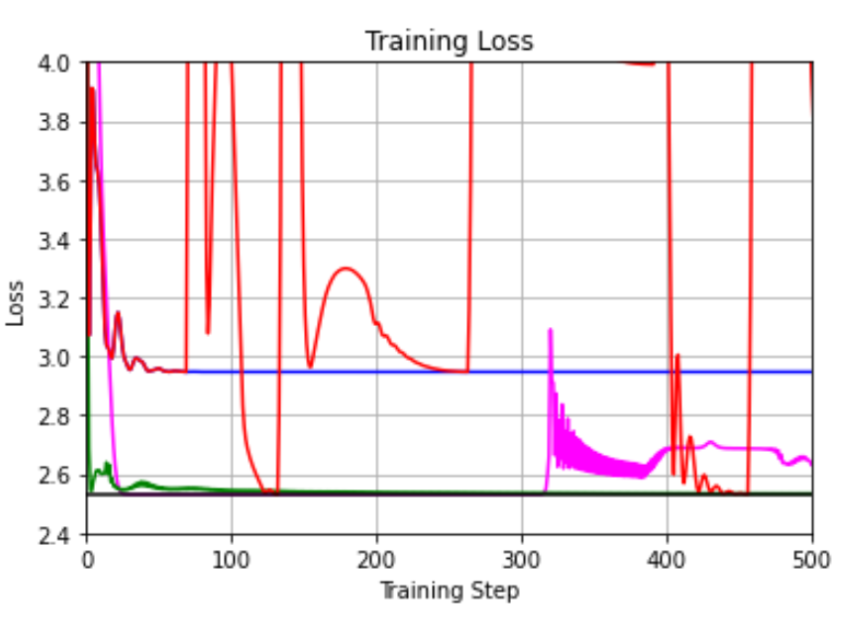}  \\

{\bfseries

Figure 4: The figure on the left shows the learning trajectories from training steps 0-200, for each of the optimizers when initializing with random seed 2534. Hill-ADAM and NADAM are the only two optimizers reaching the 2.53 error (global minimum). The figure on the right shows the learning trajectories from training steps 0-500, for each optimizer (initialization using seed 100). Again, Hill-ADAM and NADAM are the only optimizers reaching 2.53 error.

}

\includegraphics[width = 7cm, height = 7cm]{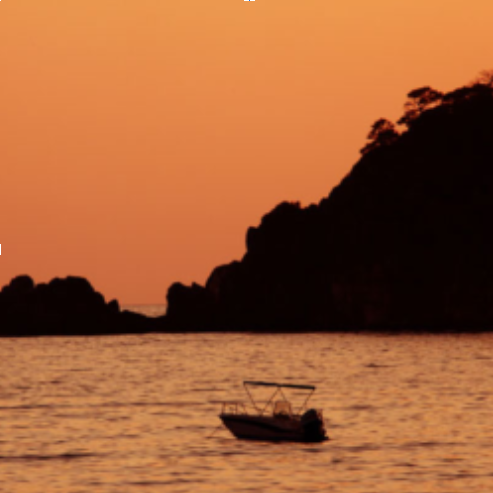}
\includegraphics[width = 7cm, height = 7cm]{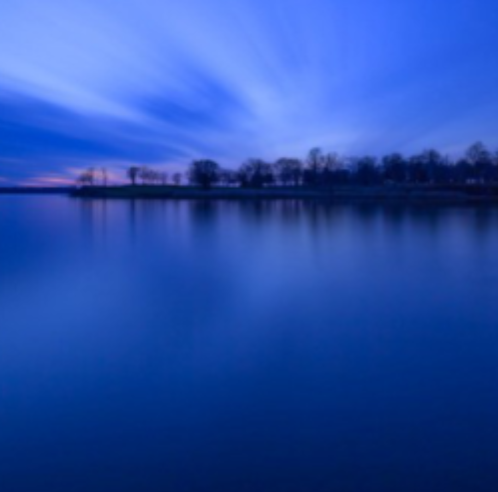} \\

{\bfseries

Figure 5: The figure on the left is the source image, and the figure on the right is the target image. The goal is to transform the yellow/orange saturated image on the left to have a cooler palette (similar to target image). Yet, the image must be slightly less blue saturated (handled by regularization). \\
 
}

\includegraphics[width = 7cm, height = 7cm]{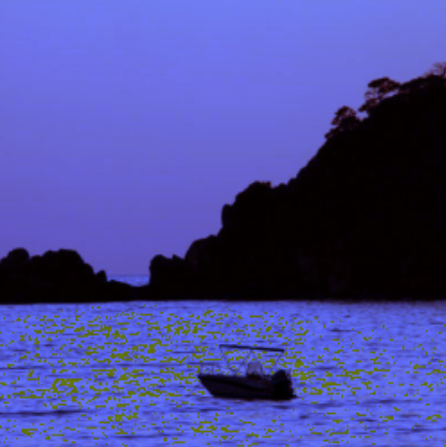}
\includegraphics[width = 7cm, height = 7cm]{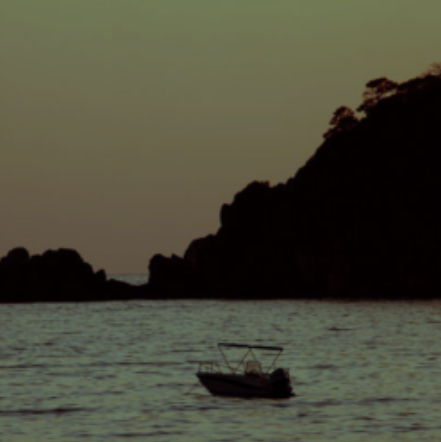} \\

{\bfseries

Figure 6: The figure on the left is the resulting, color corrected source image, based on the RGB gains found by the ADAM optimizer. The color corrected source image based on Hill-ADAM's found gains is on the right. Hill-ADAM adheres to our original goal: cooling the yellow-orange saturation while avoiding extensive blue saturation.

}

In Table 2, we tested Hill-ADAM in comparison to the benchmark ADAM optimization algorithm, along with RMSprop and NADAM algorithms. We understand that initialization can vary results significantly, so the tests were done with several initializations (summarized in the table). The table represents the training loss converged upon by each of the algorithms, for each initialization. From the table, it can be seen that Hill-ADAM converges upon the lowest minimum (global minimum) consistently compared to ADAM and RMSprop, and is competitive with NADAM.

\maketitle
\section{Discussion:}
In Table 1, as the mean approached minimum for Hill-ADAM is consistently lower than that of ADAM when possible, we can deduce that Hill-ADAM has a greater likelihood of reaching the global (or lower minimum if not global). Additionally, in each of the graphs, we can see that the ADAM optimizer seldom updates the weights after reaching the local minimum. This is shown by the stagnant training cost towards the end. Hill-ADAM avoids the stagnation by optimizing for the maximum immediately when the loss stays constant. For example, in Figure 2 (image on the left), Hill-ADAM initially lands at a loss of 5.19 (a local minimum) as shown at around training step 18150. Rather than converging and refining the loss at this point, Hill-ADAM worked to maximize the loss function and approached an error of about 15 at training step 18400. Hill-ADAM minimized the loss again and reached the global minimum of 2.88 at timestep 11950. The optimizer reached the local maximum between the two minima at training step 18400, calculating the steps needed to reach the second minimum (in this case the global minimum). The chances of escaping the local minimum that ADAM would have otherwise stayed stuck at increases significantly. The chances of finding a lower local minimum increases, \textit{even if} Hill-ADAM is unable to pinpoint the exact global minimum.   

There are special cases in which the learning rate improves the behavior of Hill-Adam. For example, if the loss function has a cubic behavior (ex. fourth order equation in Table 1), Hill-ADAM may get stuck at the point with the zero gradient. In that case Hill-ADAM will maximize, leading to a dead end, which results in a cycle in which the global minimum is never reached. This issue can be avoided by increasing the learning rate. The training steps taken will become larger, meaning the loss difference is less likely to be within epsilon. The global minimum would be found without need for maximization. 

In Table 2, we compared ADAM's performance to Hill-ADAM's, along with that of other effective optimizers such as RMSprop and NADAM. Hill-ADAM consistently outperforms ADAM and RMSprop.

From Figure 4, we can see that ADAM is stuck in the local minimum for the rest of the training because it was unable to escape the local maxima hindering it from reaching the global minimum. As a result, it oscillated around the 2.9 error (local minimum) and settled at that point. Hill-ADAM, however, upon reaching the same local minimum, immediately began the maximization process to escape and found a different, smaller minimum upon the second minimization stage. By exploring the loss space, Hill-ADAM was able to find the global minimum (2.53).

\maketitle
\section{Conclusion:}
When applied to prescribed loss spaces, existing optimization methods like the ADAM algorithm tend to converge towards the first minimum it encounters, even if it is not the global minimum. This is typically a challenging problem in machine learning because we often are not sure how much of a step (in the state space) we must take to escape the local minima, whether we use random steps or momentum. A new optimization algorithm (Hill-ADAM) is proposed given ADAM's convergence conditions, with the intent of exploring the loss space through loss minimization and maximization to increase the chance of converging towards the global minimum. The results, indeed, demonstrate Hill-ADAM's success in identifying lower minima (if not global minimum).

\maketitle
\section{Acknowledgments:}

I would like to thank Ms. My Huynh, my Multivariable Calculus/Linear Algebra instructor, for her supportive and patient mentorship in helping me develop my mathematical foundation. I would also like to thank my UC Santa Cruz Association for Computing Machinery Research Group for providing me with a kind introduction to AI/ML research procedures, corresponding mathematical application, and related fields. I am also very grateful to Dr. John Musacchio for his patience and willingness to walk me through difficult statistics concepts while making the process enjoyable. \\

\maketitle
\bibliography{references}
\bibliographystyle{plain}

\end{document}